Data and text mining

# BioBERT: a pre-trained biomedical language representation model for biomedical text mining

Jinhyuk Lee [1,†], Wonjin Yoon [1,†], Sungdong Kim [2], Donghyeon Kim [1], Sunkyu Kim [1], Chan Ho So [3] and Jaewoo Kang [1,3,*]

[1]Department of Computer Science and Engineering, Korea University, Seoul 02841, Korea, [2]Clova AI Research, Naver Corp, Seong-Nam 13561, Korea and [3]Interdisciplinary Graduate Program in Bioinformatics, Korea University, Seoul 02841, Korea

*To whom correspondence should be addressed.
†The authors wish it to be known that the first two authors contributed equally.
Associate Editor: Jonathan Wren

## Abstract

**Motivation:** Biomedical text mining is becoming increasingly important as the number of biomedical documents rapidly grows. With the progress in natural language processing (NLP), extracting valuable information from biomedical literature has gained popularity among researchers, and deep learning has boosted the development of effective biomedical text mining models. However, directly applying the advancements in NLP to biomedical text mining often yields unsatisfactory results due to a word distribution shift from general domain corpora to biomedical corpora. In this article, we investigate how the recently introduced pre-trained language model BERT can be adapted for biomedical corpora.
**Results:** We introduce BioBERT (Bidirectional Encoder Representations from Transformers for Biomedical Text Mining), which is a domain-specific language representation model pre-trained on large-scale biomedical corpora. With almost the same architecture across tasks, BioBERT largely outperforms BERT and previous state-of-the-art models in a variety of biomedical text mining tasks when pre-trained on biomedical corpora. While BERT obtains performance comparable to that of previous state-of-the-art models, BioBERT significantly outperforms them on the following three representative biomedical text mining tasks: biomedical named entity recognition (0.62% F1 score improvement), biomedical relation extraction (2.80% F1 score improvement) and biomedical question answering (12.24% MRR improvement). Our analysis results show that pre-training BERT on biomedical corpora helps it to understand complex biomedical texts.
**Availability and implementation:** We make the pre-trained weights of BioBERT freely available at https://github.com/naver/biobert-pretrained, and the source code for fine-tuning BioBERT available at https://github.com/dmis-lab/biobert.
**Contact:** kangj@korea.ac.kr

## 1 Introduction

The volume of biomedical literature continues to rapidly increase. On average, more than 3000 new articles are published every day in peer-reviewed journals, excluding pre-prints and technical reports such as clinical trial reports in various archives. PubMed alone has a total of 29M articles as of January 2019. Reports containing valuable information about new discoveries and new insights are continuously added to the already overwhelming amount of literature. Consequently, there is increasingly more demand for accurate biomedical text mining tools for extracting information from the literature.

Recent progress of biomedical text mining models was made possible by the advancements of deep learning techniques used in natural language processing (NLP). For instance, Long Short-Term Memory (LSTM) and Conditional Random Field (CRF) have greatly improved performance in biomedical named entity recognition (NER) over the last few years (Giorgi and Bader, 2018; Habibi et al., 2017; Wang et al., 2018; Yoon et al., 2019). Other deep learning based models have made improvements in biomedical text mining tasks such as relation extraction (RE) (Bhasuran and Natarajan, 2018; Lim and Kang, 2018) and question answering (QA) (Wiese et al., 2017).

However, directly applying state-of-the-art NLP methodologies to biomedical text mining has limitations. First, as recent word representation models such as Word2Vec (Mikolov et al., 2013), ELMo (Peters et al., 2018) and BERT (Devlin et al., 2019) are trained and tested mainly on datasets containing general domain texts (e.g. Wikipedia), it is difficult to estimate their performance on datasets containing biomedical texts. Also, the word distributions of general and biomedical corpora are quite different, which can often be a problem for biomedical text mining models. As a result, recent models in biomedical text mining rely largely on adapted versions of word representations (Habibi et al., 2017; Pyysalo et al., 2013).

In this study, we hypothesize that current state-of-the-art word representation models such as BERT need to be trained on biomedical corpora to be effective in biomedical text mining tasks. Previously, Word2Vec, which is one of the most widely known context independent word representation models, was trained on biomedical corpora which contain terms and expressions that are usually not included in a general domain corpus (Pyysalo et al., 2013). While ELMo and BERT have proven the effectiveness of contextualized word representations, they cannot obtain high performance on biomedical corpora because they are pre-trained on only general domain corpora. As BERT achieves very strong results on various NLP tasks while using almost the same structure across the tasks, adapting BERT for the biomedical domain could potentially benefit numerous biomedical NLP researches.

## 2 Approach

In this article, we introduce BioBERT, which is a pre-trained language representation model for the biomedical domain. The overall process of pre-training and fine-tuning BioBERT is illustrated in Figure 1. First, we initialize BioBERT with weights from BERT, which was pre-trained on general domain corpora (English Wikipedia and BooksCorpus). Then, BioBERT is pre-trained on biomedical domain corpora (PubMed abstracts and PMC full-text articles). To show the effectiveness of our approach in biomedical text mining, BioBERT is fine-tuned and evaluated on three popular biomedical text mining tasks (NER, RE and QA). We test various pre-training strategies with different combinations and sizes of general domain corpora and biomedical corpora, and analyze the effect of each corpus on pre-training. We also provide in-depth analyses of BERT and BioBERT to show the necessity of our pre-training strategies.

The contributions of our paper are as follows:

- BioBERT is the first domain-specific BERT based model pre-trained on biomedical corpora for 23 days on eight NVIDIA V100 GPUs.
- We show that pre-training BERT on biomedical corpora largely improves its performance. BioBERT obtains higher F1 scores in biomedical NER (**0.62**) and biomedical RE (**2.80**), and a higher MRR score (**12.24**) in biomedical QA than the current state-of-the-art models.
- Compared with most previous biomedical text mining models that are mainly focused on a single task such as NER or QA, our model BioBERT achieves state-of-the-art performance on various biomedical text mining tasks, while requiring only minimal architectural modifications.
- We make our pre-processed datasets, the pre-trained weights of BioBERT and the source code for fine-tuning BioBERT publicly available.

## 3 Materials and methods

BioBERT basically has the same structure as BERT. We briefly discuss the recently proposed BERT, and then we describe in detail the pre-training and fine-tuning process of BioBERT.

### 3.1 BERT: bidirectional encoder representations from transformers

Learning word representations from a large amount of unannotated text is a long-established method. While previous models (e.g. Word2Vec (Mikolov et al., 2013), GloVe (Pennington et al., 2014)) focused on learning context independent word representations, recent works have focused on learning context dependent word representations. For instance, ELMo (Peters et al., 2018) uses a bidirectional language model, while CoVe (McCann et al., 2017) uses machine translation to embed context information into word representations.

BERT (Devlin et al., 2019) is a contextualized word representation model that is based on a masked language model and pre-trained using bidirectional transformers (Vaswani et al., 2017). Due to the nature of language modeling where future words cannot be seen, previous language models were limited to a combination of two unidirectional language models (i.e. left-to-right and right-to-left). BERT uses a masked language model that predicts randomly masked words in a sequence, and hence can be used for learning bidirectional representations. Also, it obtains state-of-the-art performance on most NLP tasks, while requiring minimal task-specific architectural modification. According to the authors of BERT, incorporating information from bidirectional representations, rather than unidirectional representations, is crucial for representing words in natural language. We hypothesize that such bidirectional representations are also critical in biomedical text mining as complex relationships between biomedical terms often exist in a biomedical corpus (Krallinger et al., 2017). Due to the space limitations, we refer readers to Devlin et al. (2019) for a more detailed description of BERT.

### 3.2 Pre-training BioBERT

As a general purpose language representation model, BERT was pre-trained on English Wikipedia and BooksCorpus. However, biomedical domain texts contain a considerable number of domain-specific

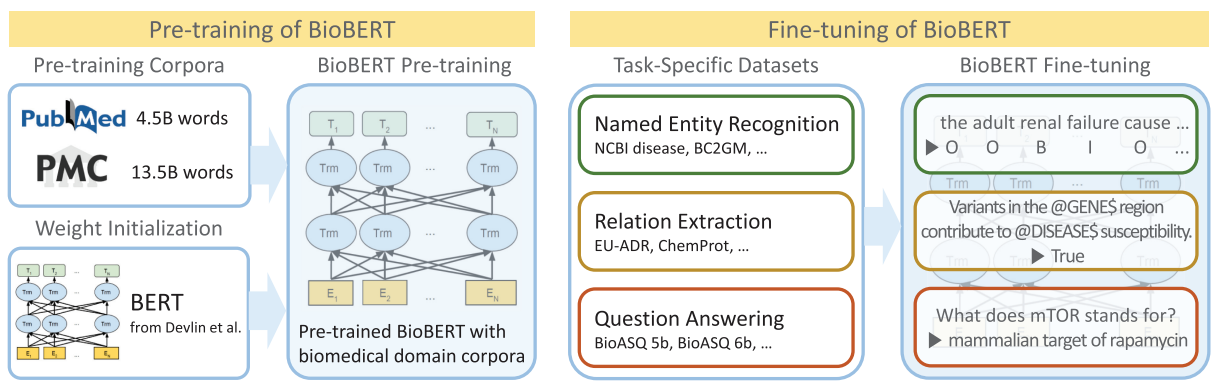

**Fig. 1.** Overview of the pre-training and fine-tuning of BioBERT

BioBERT
3BioBERT 3

**Table 1.** List of text corpora used for BioBERT

| Corpus | Number of words | Domain |
| --- | --- | --- |
| English Wikipedia | 2.5B | General |
| BooksCorpus | 0.8B | General |
| PubMed Abstracts | 4.5B | Biomedical |
| PMC Full-text articles | 13.5B | Biomedical |

proper nouns (e.g. BRCA1, c.248T>C) and terms (e.g. transcriptional, antimicrobial), which are understood mostly by biomedical researchers. As a result, NLP models designed for general purpose language understanding often obtains poor performance in biomedical text mining tasks. In this work, we pre-train BioBERT on PubMed abstracts (PubMed) and PubMed Central full-text articles (PMC). The text corpora used for pre-training of BioBERT are listed in Table 1, and the tested combinations of text corpora are listed in Table 2. For computational efficiency, whenever the Wiki + Books corpus were used for pre-training, we initialized BioBERT with the pre-trained BERT model provided by Devlin et al. (2019). We define BioBERT as a language representation model whose pre-training corpora includes biomedical corpora (e.g. BioBERT (+ PubMed)).

For tokenization, BioBERT uses WordPiece tokenization (Wu et al., 2016), which mitigates the out-of-vocabulary issue. With WordPiece tokenization, any new words can be represented by frequent subwords (e.g. *Immunoglobulin => I ##mm ##uno ##g ##lo ##bul ##in*). We found that using cased vocabulary (not lower-casing) results in slightly better performances in downstream tasks. Although we could have constructed new WordPiece vocabulary based on biomedical corpora, we used the original vocabulary of $BERT_{BASE}$ for the following reasons: (i) compatibility of BioBERT with BERT, which allows BERT pre-trained on general domain corpora to be re-used, and makes it easier to interchangeably use existing models based on BERT and BioBERT and (ii) any new words may still be represented and fine-tuned for the biomedical domain using the original WordPiece vocabulary of BERT.

### 3.3 Fine-tuning BioBERT

With minimal architectural modification, BioBERT can be applied to various downstream text mining tasks. We fine-tune BioBERT on the following three representative biomedical text mining tasks: NER, RE and QA.

*Named entity recognition* is one of the most fundamental biomedical text mining tasks, which involves recognizing numerous domain-specific proper nouns in a biomedical corpus. While most previous works were built upon different combinations of LSTMs and CRFs (Giorgi and Bader, 2018; Habibi et al., 2017; Wang et al., 2018), BERT has a simple architecture based on bidirectional transformers. BERT uses a single output layer based on the representations from its last layer to compute only token level BIO2 probabilities. Note that while previous works in biomedical NER often used word embeddings trained on PubMed or PMC corpora (Habibi et al., 2017; Yoon et al., 2019), BioBERT directly learns WordPiece embeddings during pre-training and fine-tuning. For the evaluation metrics of NER, we used entity level precision, recall and F1 score.

*Relation extraction* is a task of classifying relations of named entities in a biomedical corpus. We utilized the sentence classifier of the original version of BERT, which uses a [CLS] token for the classification of relations. Sentence classification is performed using a single output layer based on a [CLS] token representation from BERT. We anonymized target named entities in a sentence using pre-defined tags such as @GENE$ or @DISEASE$. For instance, a sentence with two target entities (gene and disease in this case) is represented as "*Serine at position 986 of @GENE$ may be an independent genetic predictor of angiographic @DISEASE$.*" The precision, recall and F1 scores on the RE task are reported.

*Question answering* is a task of answering questions posed in natural language given related passages. To fine-tune BioBERT for QA, we used the same BERT architecture used for SQuAD (Rajpurkar et al., 2016). We used the BioASQ factoid datasets because their format is similar to that of SQuAD. Token level probabilities for the start/end location of answer phrases are computed using a single output layer. However, we observed that about 30% of the BioASQ factoid questions were unanswerable in an extractive QA setting as the exact answers did not appear in the given passages. Like Wiese et al. (2017), we excluded the samples with unanswerable questions from the training sets. Also, we used the same pre-training process of Wiese et al. (2017), which uses SQuAD, and it largely improved the performance of both BERT and BioBERT. We used the following evaluation metrics from BioASQ: strict accuracy, lenient accuracy and mean reciprocal rank.

**Table 2.** Pre-training BioBERT on different combinations of the following text corpora: English Wikipedia (Wiki), BooksCorpus (Books), PubMed abstracts (PubMed) and PMC full-text articles (PMC)

| Model | Corpus combination |
| --- | --- |
| BERT (Devlin et al., 2019) | Wiki + Books |
| BioBERT (+PubMed) | Wiki + Books + PubMed |
| BioBERT (+PMC) | Wiki + Books + PMC |
| BioBERT (+PubMed + PMC) | Wiki + Books + PubMed + PMC |

## 4 Results

### 4.1 Datasets

The statistics of biomedical NER datasets are listed in Table 3. We used the pre-processed versions of all the NER datasets provided by Wang et al. (2018) except the 2010 i2b2/VA, JNLPBA and Species-800 datasets. The pre-processed NCBI Disease dataset has fewer annotations than the original dataset due to the removal of duplicate articles from its training set. We used the CoNLL format (https://github.com/spyysalo/standoff2conll) for pre-processing the 2010 i2b2/VA and JNLPBA datasets. The Species-800 dataset was pre-processed and split based on the dataset of Pyysalo (https://github.com/spyysalo/s800). We did not use alternate annotations for the BC2GM dataset, and all NER evaluations are based on entity-level exact matches. Note that although there are several other recently introduced high quality biomedical NER datasets (Mohan and Li, 2019), we use datasets that are frequently used by many biomedical NLP researchers, which makes it much easier to compare our work with theirs. The RE datasets contain gene–disease relations and protein–chemical relations (Table 4). Pre-processed GAD and EU-ADR datasets are available with our provided codes. For the CHEMPROT dataset, we used the same pre-processing procedure described in Lim and Kang (2018). We used the BioASQ factoid datasets, which can be converted into the same format as the SQuAD dataset (Table 5). We used full abstracts (PMIDs) and related questions and answers provided by the BioASQ organizers. We have made the pre-processed BioASQ datasets publicly available. For all the datasets, we used the same dataset splits used in previous works (Lim and Kang, 2018; Tsatsaronis et al., 2015; Wang et al., 2018) for a fair evaluation; however, the splits of LINAAEUS and Species-800 could not be found from Giorgi and Bader (2018) and may be different. Like previous work (Bhasuran and Natarajan, 2018), we reported the performance of 10-fold cross-validation on datasets that do not have separate test sets (e.g. GAD, EU-ADR).

We compare BERT and BioBERT with the current state-of-the-art models and report their scores. Note that the state-of-the-art models each have a different architecture and training procedure. For instance, the state-of-the-art model by Yoon et al. (2019) trained on the JNLPBA dataset is based on multiple Bi-LSTM CRF models with character level CNNs, while the state-of-the-art model by Giorgi and Bader (2018) trained on the LINNAEUS dataset uses a Bi-LSTM CRF model with character level LSTMs and is additionally trained on silver-standard datasets. On the other hand, BERT and

**Table 3.** Statistics of the biomedical named entity recognition datasets

| Dataset | Entity type | Number of annotations |
| --- | --- | --- |
| NCBI Disease (Doğan et al., 2014) | Disease | 6881 |
| 2010 i2b2/VA (Uzuner et al., 2011) | Disease | 19 665 |
| BC5CDR (Li et al., 2016) | Disease | 12 694 |
| BC5CDR (Li et al., 2016) | Drug/Chem. | 15 411 |
| BC4CHEMD (Krallinger et al., 2015) | Drug/Chem. | 79 842 |
| BC2GM (Smith et al., 2008) | Gene/Protein | 20 703 |
| JNLPBA (Kim et al., 2004) | Gene/Protein | 35 460 |
| LINNAEUS (Gerner et al., 2010) | Species | 4077 |
| Species-800 (Pafilis et al., 2013) | Species | 3708 |

*Note:* The number of annotations from Habibi et al. (2017) and Zhu et al. (2018) is provided.

**Table 4.** Statistics of the biomedical relation extraction datasets

| Dataset | Entity type | Number of relations |
| --- | --- | --- |
| GAD (Bravo et al., 2015) | Gene–disease | 5330 |
| EU-ADR (Van Mulligen et al., 2012) | Gene–disease | 355 |
| CHEMPROT (Krallinger et al., 2017) | Protein–chemical | 10 031 |

*Note:* For the CHEMPROT dataset, the number of relations in the training, validation and test sets was summed.

**Table 5.** Statistics of biomedical question answering datasets

| Dataset | Number of train | Number of test |
| --- | --- | --- |
| BioASQ 4b-factoid (Tsatsaronis et al., 2015) | 327 | 161 |
| BioASQ 5b-factoid (Tsatsaronis et al., 2015) | 486 | 150 |
| BioASQ 6b-factoid (Tsatsaronis et al., 2015) | 618 | 161 |

BioBERT have exactly the same structure, and use only the gold standard datasets and not any additional datasets.

### 4.2 Experimental setups

We used the BERT$_{BASE}$ model pre-trained on English Wikipedia and BooksCorpus for 1M steps. BioBERT v1.0 (+ PubMed + PMC) is the version of BioBERT (+ PubMed + PMC) trained for 470 K steps. When using both the PubMed and PMC corpora, we found that 200K and 270K pre-training steps were optimal for PubMed and PMC, respectively. We also used the ablated versions of BioBERT v1.0, which were pre-trained on only PubMed for 200K steps (BioBERT v1.0 (+ PubMed)) and PMC for 270K steps (BioBERT v1.0 (+ PMC)). After our initial release of BioBERT v1.0, we pre-trained BioBERT on PubMed for 1M steps, and we refer to this version as BioBERT v1.1 (+ PubMed). Other hyper-parameters such as batch size and learning rate scheduling for pre-training BioBERT are the same as those for pre-training BERT unless stated otherwise.

We pre-trained BioBERT using Naver Smart Machine Learning (NSML) (Sung et al., 2017), which is utilized for large-scale experiments that need to be run on several GPUs. We used eight NVIDIA V100 (32GB) GPUs for the pre-training. The maximum sequence length was fixed to 512 and the mini-batch size was set to 192, resulting in 98 304 words per iteration. It takes more than 10 days to pre-train BioBERT v1.0 (+ PubMed + PMC) nearly 23 days for BioBERT v1.1 (+ PubMed) in this setting. Despite our best efforts to use BERT$_{LARGE}$, we used only BERT$_{BASE}$ due to the computational complexity of BERT$_{LARGE}$.

We used a single NVIDIA Titan Xp (12GB) GPU to fine-tune BioBERT on each task. Note that the fine-tuning process is more computationally efficient than pre-training BioBERT. For fine-tuning, a batch size of 10, 16, 32 or 64 was selected, and a learning rate of 5e−5, 3e−5 or 1e−5 was selected. Fine-tuning BioBERT on QA and RE tasks took less than an hour as the size of the training data is much smaller than that of the training data used by Devlin et al. (2019). On the other hand, it takes more than 20 epochs for BioBERT to reach its highest performance on the NER datasets.

### 4.3 Experimental results

The results of NER are shown in Table 6. First, we observe that BERT, which was pre-trained on only the general domain corpus is quite effective, but the micro averaged F1 score of BERT was lower (2.01 lower) than that of the state-of-the-art models. On the other hand, BioBERT achieves higher scores than BERT on all the datasets. BioBERT outperformed the state-of-the-art models on six out of nine datasets, and BioBERT v1.1 (+ PubMed) outperformed the state-of-the-art models by 0.62 in terms of micro averaged F1 score. The relatively low scores on the LINNAEUS dataset can be attributed to the following: (i) the lack of a silver-standard dataset for training previous state-of-the-art models and (ii) different training/test set splits used in previous work (Giorgi and Bader, 2018), which were unavailable.

The RE results of each model are shown in Table 7. BERT achieved better performance than the state-of-the-art model on the CHEMPROT dataset, which demonstrates its effectiveness in RE. On average (micro), BioBERT v1.0 (+ PubMed) obtained a higher F1 score (2.80 higher) than the state-of-the-art models. Also, BioBERT achieved the highest F1 scores on 2 out of 3 biomedical datasets.

The QA results are shown in Table 8. We micro averaged the best scores of the state-of-the-art models from each batch. BERT obtained a higher micro averaged MRR score (7.0 higher) than the state-of-the-art models. All versions of BioBERT significantly outperformed BERT and the state-of-the-art models, and in particular, BioBERT v1.1 (+ PubMed) obtained a strict accuracy of 38.77, a lenient accuracy of 53.81 and a mean reciprocal rank score of 44.77, all of which were micro averaged. On all the biomedical QA datasets, BioBERT achieved new state-of-the-art performance in terms of MRR.

## 5 Discussion

We used additional corpora of different sizes for pre-training and investigated their effect on performance. For BioBERT v1.0 (+ PubMed), we set the number of pre-training steps to 200K and varied the size of the PubMed corpus. Figure 2(a) shows that the performance of BioBERT v1.0 (+ PubMed) on three NER datasets (NCBI Disease, BC2GM, BC4CHEMD) changes in relation to the size of the PubMed corpus. Pre-training on 1 billion words is quite effective, and the performance on each dataset mostly improves until 4.5 billion words. We also saved the pre-trained weights from BioBERT (+ PubMed) at different pre-training steps to measure how the number of pre-training steps affects its performance on fine-tuning tasks. Figure 2(b) shows the performance changes of BioBERT v1.0 (+ PubMed) on the same three NER datasets in relation to the number of pre-training steps. The results clearly show that the performance on each dataset improves as the number of pre-training steps increases. Finally, Figure 2(c) shows the absolute performance improvements of BioBERT v1.0 (+ PubMed + PMC) over BERT on all 15 datasets. F1 scores were used for NER/RE, and MRR scores were used for QA. BioBERT significantly improves performance on most of the datasets.

As shown in Table 9, we sampled predictions from BERT and BioBERT v1.1 (+PubMed) to see the effect of pre-training on downstream tasks. BioBERT can recognize biomedical named entities that BERT cannot and can find the exact boundaries of named

**Table 6.** Test results in biomedical named entity recognition

| Type | Datasets | Metrics | SOTA | BERT (Wiki + Books) | BioBERT v1.0 (+ PubMed) | BioBERT v1.0 (+ PMC) | BioBERT v1.0 (+ PubMed + PMC) | BioBERT v1.1 (+ PubMed) |
|---|---|---|---|---|---|---|---|---|
| Disease | NCBI disease | P | 88.30 | 84.12 | 86.76 | 86.16 | **89.04** | 88.22 |
| | | R | 89.00 | 87.19 | 88.02 | 89.48 | 89.69 | **91.25** |
| | | F | 88.60 | 85.63 | 87.38 | 87.79 | 89.36 | **89.71** |
| | 2010 i2b2/VA | P | 87.44 | 84.04 | 85.37 | 85.55 | **87.50** | 86.93 |
| | | R | 86.25 | 84.08 | 85.64 | 85.72 | 85.44 | **86.53** |
| | | F | **86.84** | 84.06 | 85.51 | 85.64 | 86.46 | 86.73 |
| | BC5CDR | P | **89.61** | 81.97 | 85.80 | 84.67 | 85.86 | 86.47 |
| | | R | 83.09 | 82.48 | 86.60 | 85.87 | 87.27 | **87.84** |
| | | F | 86.23 | 82.41 | 86.20 | 85.27 | 86.56 | **87.15** |
| Drug/chem. | BC5CDR | P | **94.26** | 90.94 | 92.52 | 92.46 | 93.27 | 93.68 |
| | | R | 92.38 | 91.38 | 92.76 | 92.63 | **93.61** | 93.26 |
| | | F | 93.31 | 91.16 | 92.64 | 92.54 | 93.44 | **93.47** |
| | BC4CHEMD | P | 92.29 | 91.19 | 91.77 | 91.65 | 92.23 | **92.80** |
| | | R | 90.01 | 88.92 | 90.77 | 90.30 | 90.61 | **91.92** |
| | | F | 91.14 | 90.04 | 91.26 | 90.97 | 91.41 | **92.36** |
| Gene/protein | BC2GM | P | 81.81 | 81.17 | 81.72 | 82.86 | **85.16** | 84.32 |
| | | R | 81.57 | 82.42 | 83.38 | 84.21 | 83.65 | **85.12** |
| | | F | 81.69 | 81.79 | 82.54 | 83.53 | 84.40 | **84.72** |
| | JNLPBA | P | **74.43** | 69.57 | 71.11 | 71.17 | 72.68 | 72.24 |
| | | R | 83.22 | 81.20 | 83.11 | 82.76 | 83.21 | **83.56** |
| | | F | **78.58** | 74.94 | 76.65 | 76.53 | 77.59 | 77.49 |
| Species | LINNAEUS | P | 92.80 | 91.17 | 91.83 | 91.62 | **93.84** | 90.77 |
| | | R | **94.29** | 84.30 | 84.72 | 85.48 | 86.11 | 85.83 |
| | | F | **93.54** | 87.60 | 88.13 | 88.45 | 89.81 | 88.24 |
| | Species-800 | P | **74.34** | 69.35 | 70.60 | 71.54 | 72.84 | 72.80 |
| | | R | 75.96 | 74.05 | 75.75 | 74.71 | **77.97** | 75.36 |
| | | F | 74.98 | 71.63 | 73.08 | 73.09 | **75.31** | 74.06 |

*Notes:* Precision (P), Recall (R) and F1 (F) scores on each dataset are reported. The best scores are in bold, and the second best scores are underlined. We list the scores of the state-of-the-art (SOTA) models on different datasets as follows: scores of Xu *et al.* (2019) on NCBI Disease, scores of Sachan *et al.* (2018) on BC2GM, scores of Zhu *et al.* (2018) (single model) on 2010 i2b2/VA, scores of Lou *et al.* (2017) on BC5CDR-disease, scores of Luo *et al.* (2018) on BC4CHEMD, scores of Yoon *et al.* (2019) on BC5CDR-chemical and JNLPBA and scores of Giorgi and Bader (2018) on LINNAEUS and Species-800.

**Table 7.** Biomedical relation extraction test results

| Relation | Datasets | Metrics | SOTA | BERT (Wiki + Books) | BioBERT v1.0 (+ PubMed) | BioBERT v1.0 (+ PMC) | BioBERT v1.0 (+ PubMed + PMC) | BioBERT v1.1 (+ PubMed) |
|---|---|---|---|---|---|---|---|---|
| Gene–disease | GAD | P | **79.21** | 74.28 | 76.43 | 75.20 | 75.95 | 77.32 |
| | | R | **89.25** | 85.11 | 87.65 | 86.15 | 88.08 | 82.68 |
| | | F | **83.93** | 79.29 | 81.61 | 80.24 | 81.52 | 79.83 |
| | EU-ADR | P | 76.43 | 75.45 | 78.04 | **81.05** | 80.92 | 77.86 |
| | | R | **98.01** | 96.55 | 93.86 | 93.90 | 90.81 | 83.55 |
| | | F | 85.34 | 84.62 | 84.44 | **86.51** | 84.83 | 79.74 |
| Protein–chemical | CHEMPROT | P | 74.80 | 76.02 | 76.05 | **77.46** | 75.20 | 77.02 |
| | | R | 56.00 | 71.60 | 74.33 | 72.94 | 75.09 | **75.90** |
| | | F | 64.10 | 73.74 | 75.18 | 75.13 | 75.14 | **76.46** |

*Notes:* Precision (P), Recall (R) and F1 (F) scores on each dataset are reported. The best scores are in bold, and the second best scores are underlined. The scores on GAD and EU-ADR were obtained from Bhasuran and Natarajan (2018), and the scores on CHEMPROT were obtained from Lim and Kang (2018).

entities. While BERT often gives incorrect answers to simple biomedical questions, BioBERT provides correct answers to such questions. Also, BioBERT can provide longer named entities as answers.

## 6 Conclusion

In this article, we introduced BioBERT, which is a pre-trained language representation model for biomedical text mining. We showed that pre-training BERT on biomedical corpora is crucial in applying it to the biomedical domain. Requiring minimal task-specific architectural modification, BioBERT outperforms previous models on biomedical text mining tasks such as NER, RE and QA.

The pre-released version of BioBERT (January 2019) has already been shown to be very effective in many biomedical text mining tasks such as NER for clinical notes (Alsentzer *et al.*, 2019), human phenotype-gene RE (Sousa *et al.*, 2019) and clinical temporal RE (Lin *et al.*, 2019). The following updated versions of BioBERT will be available to the bioNLP community: (i) BioBERT$_{BASE}$ and BioBERT$_{LARGE}$ trained on only PubMed abstracts without initialization from the existing BERT model and (ii) BioBERT$_{BASE}$ and BioBERT$_{LARGE}$ trained on domain-specific vocabulary based on WordPiece.

**Table 8.** Biomedical question answering test results

| Datasets | Metrics | SOTA | BERT (Wiki + Books) | BioBERT v1.0 (+ PubMed) | (+ PMC) | (+ PubMed + PMC) | BioBERT v1.1 (+ PubMed) |
|---|---|---|---|---|---|---|---|
| BioASQ 4b | S | 20.01 | 27.33 | 25.47 | 26.09 | **28.57** | 27.95 |
|  | L | 28.81 | 44.72 | 44.72 | 42.24 | **47.82** | 44.10 |
|  | M | 23.52 | 33.77 | 33.28 | 32.42 | **35.17** | 34.72 |
| BioASQ 5b | S | 41.33 | 39.33 | 41.33 | 42.00 | 44.00 | **46.00** |
|  | L | 56.67 | 52.67 | 55.33 | 54.67 | 56.67 | **60.00** |
|  | M | 47.24 | 44.27 | 46.73 | 46.93 | 49.38 | **51.64** |
| BioASQ 6b | S | 24.22 | 33.54 | **43.48** | 41.61 | 40.37 | 42.86 |
|  | L | 37.89 | 51.55 | 55.90 | 55.28 | **57.77** | **57.77** |
|  | M | 27.84 | 40.88 | 48.11 | 47.02 | 47.48 | **48.43** |

*Notes:* Strict Accuracy (S), Lenient Accuracy (L) and Mean Reciprocal Rank (M) scores on each dataset are reported. The best scores are in bold, and the second best scores are underlined. The best BioASQ 4b/5b/6b scores were obtained from the BioASQ leaderboard (http://participants-area.bioasq.org).

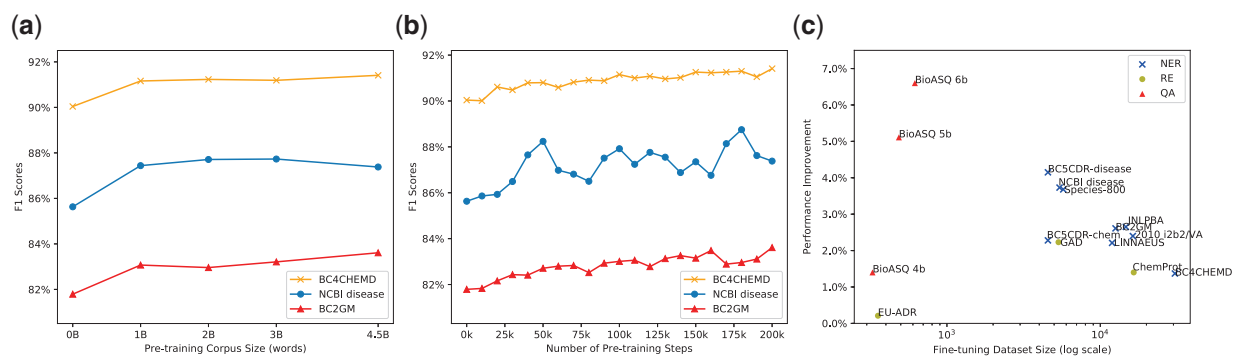

**Fig. 2.** (**a**) Effects of varying the size of the PubMed corpus for pre-training. (**b**) NER performance of BioBERT at different checkpoints. (**c**) Performance improvement of BioBERT v1.0 (+ PubMed + PMC) over BERT

**Table 9.** Prediction samples from BERT and BioBERT on NER and QA datasets

| Task | Dataset | Model | Sample |
|---|---|---|---|
| NER | NCBI disease | BERT | WT1 missense mutations, associated with male pseudohermaphroditism in **Denys–Drash syndrome**, fail to ... |
|  |  | BioBERT | WT1 missense mutations, associated with **male pseudohermaphroditism** in **Denys–Drash syndrome**, fail to ... |
|  | BC5CDR (Drug/Chem.) | BERT | ... a case of oral **penicillin anaphylaxis** is described, and the terminology ... |
|  |  | BioBERT | ... a case of oral **penicillin** anaphylaxis is described, and the terminology ... |
|  | BC2GM | BERT | Like the DMA, but unlike all other mammalian class II A genes, the zebrafish gene codes for two cysteine residues ... |
|  |  | BioBERT | Like the **DMA**, but unlike all other mammalian class II A genes, the zebrafish gene codes for two cysteine residues ... |
| QA | BioASQ 6b-factoid |  | Q: Which type of urinary incontinence is diagnosed with the Q tip test? |
|  |  | BERT | A total of 25 women affected by clinical **stress** urinary incontinence (SUI) were enrolled. After undergoing (...) Q-tip test, ... |
|  |  | BioBERT | A total of 25 women affected by clinical **stress urinary incontinence** (SUI) were enrolled. After undergoing (...) Q-tip test, ... |
|  |  |  | Q: Which bacteria causes erythrasma? |
|  |  | BERT | **Corynebacterium** minutissimum is the bacteria that leads to cutaneous eruptions of erythrasma ... |
|  |  | BioBERT | **Corynebacterium minutissimum** is the bacteria that leads to cutaneous eruptions of erythrasma ... |

*Note:* Predicted named entities for NER and predicted answers for QA are in bold.

## Funding

This research was supported by the National Research Foundation of Korea(NRF) funded by the Korea government (NRF-2017R1A2A1A17069645, NRF-2017M3C4A7065887, NRF-2014M3C9A3063541).